# A SOFT COMPUTING MODEL FOR PHYSICIANS' DECISION PROCESS

Siddhartha Sankar Biswas

Abstract— In this paper the author presents a kind of Soft Computing Technique, mainly an application of fuzzy set theory of Prof. Zadeh [16], on a problem of Medical Experts Systems. The choosen problem is on design of a physician's decision model which can take crisp as well as fuzzy data as input, unlike the traditional models. The author presents a mathematical model based on fuzzy set theory for physician aided evaluation of a complete representation of information emanating from the initial interview including patient past history, present symptoms, and signs observed upon physical examination and results of clinical and diagnostic tests.

——————————— ◆ ———————————

## 1 INTRODUCTION

Most of the existing mathematical models of physician decision process offered to date, especially those relative to diagnosis and patient treatment [1-3,5,9-11,13,14] suffer from the inability to incorporative all useful real data on the patient. Pertinent information so neglected or poorly modeled relate to variables that are intrinsically fuzzy but which describe that patient's health status. Considering the real situation of illiterate farmers, villages of remote rural areas on health care and medical systems, it is felt that a soft-computing based solution would be an improved solution.

### 1.1 WHAT IS SOFT COMPUTING?

Soft computing differs from conventional (hard) computing in that, (unlike hard computing), it computes exploiting the tolerance of imprecision, uncertainty, partial truth, and approximation. In effect, the role model for soft computing is the human mind. The guiding principle of soft computing is: Exploit the tolerance for imprecision, uncertainty, partial truth, and approximation to achieve tractability, robustness and low solution cost.

The principal constituents of Soft Computing (SC) are [4,6-8,12,15-17]:-
Fuzzy Logic (FL), Rough Logic, Soft Set Logic, Neural Computing (NC), Genetic Computing, Evolutionary Computation (EC) Machine Learning (ML) and Probabilistic Reasoning (PR), with the latter subsuming belief networks, Fractal Theory, Chaos Theory, Probability Theory, Possibility Theory, and Learning Theory.

### 1.2 JUSTIFICATION OF THE NEED OF SOFT COMPUTING TECHNIQUES

In this paper I present a model which is based on real life input data like patient past history, present symptoms, and signs observed upon physical examination and results of clinical and diagnostic tests. This real input data are not all crisp in nature, rather fuzzy. Most of the data are nonnumeric, viz. "good", "very good", "low pH", "high turbidity", "less alkalinity", "high BOD", "poor sanitation", etc. to list a few only out of infinity. Such type of data are fuzzy in nature. Evaluation of many objects is not possible by hard computing but by the application of powerful softcomputing techniques.

## 2 THE GENERAL DIAGNOSTIC PROCESS: A SYSTEMS DESCRIPTION

The diagnostic decision process is a sequence of decisions made by a physician in an attempt to identify and explain the ailments, disorders and diseases present in a particular sick patient. This process involves the acceptance of the patient into the physician's care, and the collection and evaluation of pertinent information at various intermittent stages. Such information, obtained through discussion, observation, and tests, is significant to the convergence upon effective preliminary and final diagnosis. Fig I illustrates this process as well as the general treatment decision process.

————————————————

• *Siddharta Sankar Biswas is with Gurgaon Institute of Technology & management, Gurgaon.*

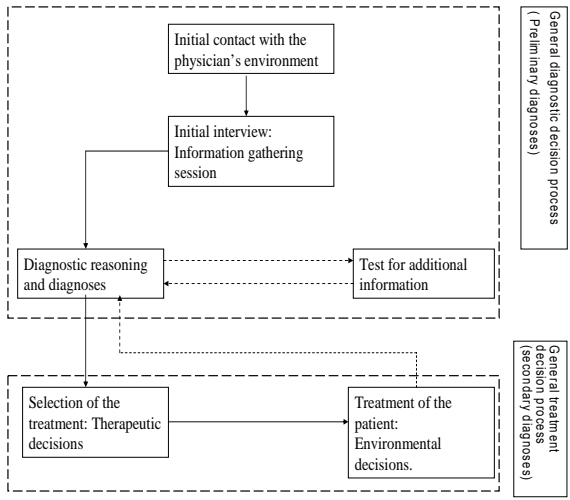

Fig 1 : System diagram of the general diagnostic and treatment processes

## 3. MATHEMATICAL MODELS: PATIENT INFORMATION

Patient information under consideration, here, can be classified into one on of the following categories:
(i)     Patient Past History.
(ii)    Present Symptoms.
(iii)   Signs Observed upon physical examination, and
(iv)    Results of clinical and diagnostic tests.

### 3.1    PATIENT PAST HISTORY

In this paper, an attempt is made to represent directly, past history with more clarity and detail, in an effort to remedy the incomplete use of information characteristic of classical techniques. Let N be the set of diseases under consideration for the diagnoses. There exists a finite set, $\Omega$, of prominent history aspects, similar to those mentioned above. This set includes non-medical as well as medical aspects which are physician designated diseases or disorders. Let N be the set of diseases under consideration for the diagnoses. For each disease $i \in N$, there exist a finite set $\Omega_i$ of prominent history aspects similar to those mentioned above. This set includes non-medical as well as medical aspects which are physician designated diseases or disorders.
           Let    $\Omega = \cup \Omega_i$  = the set of all prominent history aspects…...(1)
Let n is the cardinality of N. This set $\Omega$ has as its elements all aspects of patient history relevant to the diagnoses.  Each of these aspects or elements of $\Omega$ is either present or absent form a particular patient's history and is assumed to be binary. To represent this concept, an 1x m matrix H is created for each patient, such that each element of this matrix relates the presence or absence of the corresponding element of the set $\Omega$.
Thus   H = [ h(1), h(2), ………, h(m) ]……………………………………(2)
with h(i) $\in$ I, where I = {0,1}, and m is the cardinality of the set $\Omega$.

To mathematically convey this, let $\Omega A \subseteq \Omega$ Such that the p elements of $\Omega A$ correspond to the p diseases or disorders which might be missed or go undiagnosed, where p ≤ m.  Consider the p elements of $\Omega A$ and their corresponding elements in the matrix H.  Since both the set and matrix are finite, the matrix H can be ordered or restructured so that its first p entries correspond to the elements of $\Omega A$.

Thus

H =[ h(1), h(2),…, h(p – 1), h(p), h(p + 1),..h(m) ]. ……(3)
From this matrix H,  the two following submatrices HA and HB can be formed such that H =[ HA , HB ] ,            where
HA    =   [ h(1), h(2),….. … , h(p) ].
and
HB = [ h(p + 1),  h(p + 2),….. … , h(m) ]…….. ( 4 )

The presence or absence of each history in $\Omega A$, represented in the binary matrix HA,   might then be determined  from past, undiagnosed symptoms  and sicknesses. Since not all these past symptoms will be recalled by the patient, the physician only considers the 'prominent' symptoms of past sicknesses.
Let  Bj be the set of possible symptoms for the past undiagnosed disease j,  where 1 ≤ j ≤ p.
The fuzzy set $\theta j$ containing the 'prominent' symptoms of past disease j is defined as   below :
$\theta j$   =    { ( xij , $\mu \theta j$ (xij) ) :   xij $\in$ Bj , and  $\mu \theta j$ (xij) ≥ $\alpha$ } ,………….(5)
where   $\mu \theta j$ (xij) $\in$ [0,1]    and    0 ≤ $\alpha$ ≤ 1.

Whenever the physician designates the membership function  $\mu \theta j$ (xij) of symptom xij for disease j over a specified level $\alpha$,  the symptom of Bj becomes a 'prominent' symptom.
For each fuzzy set $\theta j$ ,   where 1 ≤ j ≤ p,      let us construct a matrix V(j),   where the elements of V(j), represent the presence or absence of xij , for every xi. belonging to $\theta j$
Thus V(j)=[v(1,j),v(2,j), ……., v(ki,j)]………….(6)
where   v(i,j) $\in$ I, I = {0,1},   i = 1, 2, 3,…., kj , 1 ≤ j ≤ p,   and kj is the cardinality of $\theta j$.
The representation v(i,j)  for symptom i of disease j is binary  and thus  assumed to lack severity levels since the specifics needed for severity determination are forgotten or altered with the passing of time.  If disease j is not designated in the patient's history,  but symptoms of disease j have existed in the patient's past, then the matrix V(j) must be determined. This matrix V(j) is then used to complete the patient history matrix H. for a specific patient and undiagnosed disease of disorder

$$f(V(j))V(j) \longrightarrow h(j) \quad \ldots\ldots\ldots\ldots\ldots\ldots(7)$$

such that $f(V(j)) \in I$, $I = \{0, 1\}$ and $1 \leq j \leq p$. Therefore, a vector of symptoms V(j) is mapped (non-fuzzy) into the jth element of HA such that h(j) equals 0 or 1. The function f(V(j)) may be very simplistic in nature, or be similar to the function presented as the general diagnosis decision model. Using this mapping the past history matrix HA can be quantified for all possible elements h(j), $1 \leq j \leq p$ and is used later in developing the various diagnoses decision models.

## 3.2 PRESENT SYMPTOMS

Classical models begin the decision process with the symptoms already designated or severity levels determined. Let $\pi_i$ = the set of possible problems that might be observed or experienced by a patient with disease i, where $i \in N$. Further, let

$$\pi = U \pi_i, i = 1, 2, 3, \ldots, n \quad \ldots\ldots\ldots(8)$$

where n is the cardinality of N.

We note that π comprises the patient related problems that are encountered for all diseases under consideration. The elements of this set are not medically designated symptoms, but are patient descriptions such as dizziness, chest pain, and inability to breathe properly. Associated with each of these problems, q, (where $q \in \pi$), is a set of factors Δq. which are important to the medical designation and severity determination of problem q. Each problem profile set Δq may have the following subsets of discrete information :

$$\Delta q_1 = \{\text{location of problem q}\}$$
$$\Delta q_2 = \{\text{longevity of problem q}\}$$
$$\Delta q_3 = \{\text{continuity of problem q}\} \quad \ldots\ldots\ldots\ldots(9)$$
$$\Delta q_4 = \{\text{defining aspects of intermittent problem q}\}$$
$$\Delta q_5 = \{\text{specifics for severity determination of problem q : fuzzy descriptions}\}.$$

where $\Delta q = \Delta q_1 U \Delta q_2 U \Delta q_3 U \Delta q_4 U \Delta q_5$ ……………………………(10)

Let $\Delta = U \Delta q$, $q = 1, 2, 3, \ldots, r$
where r is the cardinality of the set π.
If β is the collective set of medically designated symptoms for the disease under consideration, then

$$f(\pi, \Delta)\pi \longrightarrow \beta \quad \ldots\ldots\ldots\ldots\ldots\ldots\ldots(11)$$

Define a matrix B containing information obtained from the patient with regard to his problems π, and problem profiles Δq as B.

$$B = \text{matrix } (b(i,j)), \ldots\ldots\ldots\ldots\ldots\ldots\ldots(12)$$

where $i = 1, 2, 3, \ldots, s$ and $j = 1, 2, 3, \ldots, r$.

Here for a cell-element b(i, j) of this matrix B, $j=1,2,3,\ldots,r$, and

$$\ldots\ldots\ldots\ldots\ldots\ldots\ldots\ldots\ldots\ldots(13)$$

$i \in I$ is the index for characteristics of the problem profile, and $r = \max I$.

For example, the ith row of this matrix might correspond to the location of problems j, and the jth column to the profile factors in the set Δq with j = q.

Clearly, the matrix B contains all the pertinent information obtained from the verbal physician – patient interaction during the initial interview. Once the information for B has been gained, the physician must medically designate the symptoms as well as the severity levels. From the set β of all possible medically designated symptoms under consideration, let us construct a matrix A such that

$$A = [a(1), a(2), a(3), \ldots, a(t)] \ldots\ldots\ldots\ldots(14)$$

in which $a(i) \in [0,1]$ and t is the cardinality of β. The variable a(i) of the matrix A represents the severity of symptom i, with $i \in \beta$. If a symptom i is assumed to be dichotomous for the diseases under consideration then $a(i) \in \{0, 1\} \ldots\ldots\ldots(15)$
with

a(i) = 0, if symptom i is not related by the patient.
     = 1, otherwise.

Otherwise, the severity of the symptom is considered to be pertinent information and can be represented as $a(i) \in [0,1]$. As a(i) approaches 1, the severity of symptom i increases while the severity of symptom i decreases as a(i) approaches 0.

The mapping of patient related problems to medically designated symptoms can now be more precisely written in matrix from as

$$\delta (B) B \longrightarrow A \quad \ldots\ldots\ldots\ldots\ldots\ldots\ldots(16)$$

In this matrix form, a given problem j may be mapped into one more entries of A. Let b(j), a column vector of matrix B, represent the problem j and its problem profile set Δj. Then.

$$\delta (b(j)) b(j) \longrightarrow a(i) \quad \ldots\ldots\ldots\ldots(17)$$

where $j = 1, 2, \ldots r$; and $i = 1, 2, \ldots, t$.

Clearly, a(i) is binary, i.e. the map $\delta (b(j)) \in \{0, 1\}$. For this case, the function δ is very simplistic and usually incorporates the presence or absence of a few factors, primarily those of location, in the problem profile. It is obvious that if c be the relative measure or description of the symptom severity, A be the fuzzy set evaluating severity descriptions.

Then

$$\mu A(c) = b(i,j), \text{ iff } a(i) \text{ does not belong to } [0,1]. \ldots\ldots(18)$$

If a(i) is not assumed to be binary, then more complex mappings involving fuzzy set theory exist. In this case, a component of b(j), say b(i,j), often represents the membership function for a fuzzy description of the severity of problem j. The membership function reflect the painfulness of or blueness of symptoms such as headache and cyanosis. Fuzzy sets such as these are pertinent to the determination of symptom severity levels. Thus for this problem case, the function of the mapping in Eq. (17) becomes

δ (b(j)) = δ [ b(1,j), b(2,j), ……, b(k,j) ] r ………(19)

This mapping can thus involve functions of fuzzy and non-fuzzy sets. Many times, only the membership function of a fuzzy set, μA(c), is needed to determine the severity of a symptom. In such cases the eq.(19) reduces to δ(b(j)) = μA(c), ……………………………………(20)

or δ (b(j)) = g(μA(c))

where g is a simple function of fuzzy sets and other problem factors. Thus fuzzy set theory is highly useful in characterizing the severity levels of non-binary symptoms.

### 3.3 SIGNS OBSERVED UPON PHYSICAL EXAMINATION

Most models in the literature [ex. 1-3,5,9-11,13,14] either fail to include a description for severity level or simply equated symptoms with signs. We attempt to correct this deficiency by more precisely modeling varying levels of sign severity.

For the diseases under consideration, let

Φj = {signs of disease j},j ∈ N …………(21)

and Φ = U Φj where j = 1, 2, 3,………n

In matrix form, let

S = [ s(1), s(2),… . . ., s(f)] …………………(22)

where f is the cardinality of Φ, and s(j) ∈ [0, 1], j = 1, 2, . . . f.

Thus, s(j) reflects the severity of sign j in the patient. If the sign corresponding to s(j) is assumed to be binary, then s(j) ∈ {0, 1}.

Let D denotes a matrix of observables, in which

$$D = \begin{bmatrix} d(1,1) & d(1,2) & \ldots & d(1,f) \\ d(2,1) & d(2,2) & \ldots & d(2,f) \\ \ldots & & & \\ \ldots & & & \\ d(e,1) & d(e,2) & \ldots & d(e,f) \end{bmatrix}$$

with e is the number of elements in largest set of observables. An entry of the matrix is denoted by d (i,j) ……………….(24) with i ∈ I (index for each set of observables), and j = 1, 2, ……… f. Each column of matrix D, say d(j), corresponds to the set of observables for sign j. The severity of sign j, is represented thus

Ψ (b(j)) d(j) ——> s(j) ……………………………(25)

This mapping, as was the case in determining symptom severity, may or may not involve fuzzy sets. The function Ψ is usually very similar in nature to the function δ used to map patient information into symptoms in Eq. (17). Nonfuzzy mappings exist when only location factors are pertinent to severity determination. If other factors exist, they often involve the use of fuzzy sets. For example, severity aspects of a systolic heart murmur may involve fuzzy sets with regard to the loudness and quality of the sound. Specific functional relationships of Ψ(d(j)) are omitted since the general structures are similar to those presented in Eqs. (19) and (20).

### 3.4 RESULTS OF CLINICAL AND DIAGNOSTIC TESTS

Consider the set Tt of possible test results of test t. The elements of this set Tt may be discrete or continuous. Fuzzy set theory is introduced in an attempt to transform these test results in to a proper perspective and scale. For each test t, a fuzzy set Γt is created to represent the 'abnormality' of the possible test results. Tt. The membership function, reflection the degree of abnormality, must be determined for each set Γt   t = 1, 2, 3,…..k., where k equals the number of tests preformed on the patient. Let rt ∈ Tt. The degree of abnormality of this test result is reflected by the membership function μΓt ( rt) where

μΓt ( rt) rt ——> z(t) ……………………………(26)

and z(t) ∈ [0,1].

For the k tests performed on a given patient, let

Z = [ z(1) z(2) . . . z(k) ] ………………(27)

represent the test results evaluated via fuzzy set theory. Any test result can be mapped into the [0, 1] continuum using this theory, so k is the number of tests performed. The degree of membership is thus a function of a single variable, where

$$\mu\Gamma t(rt) = \begin{cases} 0, & \text{for } rt < 260 \\ \dfrac{rt}{340} - \dfrac{26}{34} & \text{for } 260 \leq rt \leq 600 \\ 1 & \text{for } rt > 600 \end{cases}$$

…………………...(28)

The result of test t, rt may possess defining aspects so that

rt ={ r1, r2,……….rtn } ………………………..(29)

where each $r_j$ for $j = 1, 2, 3,\ldots,n$ is a defining aspect. Thus, the fuzzy mapping in Eq. (26) becomes

$$r_j \longrightarrow z(t) \quad \ldots\ldots\ldots\ldots\ldots\ldots(30)$$

where $\mu_j$ is a membership function mapping $r_{t1}, r_{t2}, \ldots, r_{tn}$ into $[0, 1]$. Since these aspects $r_{tj}$, $j = 1,2,\ldots n$ are very specific to the individual test t, generalities concerning the possible structure of $\mu\Gamma(r_{t1}, r_{t2},\ldots, r_{tn})$ are difficult to make Consequently this will be investigated in a future effort.

## 4 CONCLUSION

The foregoing models are aimed at developing more complete models that are useful for the development of a reliable diagnosis decision model. Important information nets usually assumed away in previous mathematical models are explicitly considered and modeled via fuzzy set theory.

## References

[1] A.O. Esogbue, V. Aggrwal and D. Kajaulgi, Computer aided anesthesia administration, Int. J. Bio- Med. Comput. 7 (1976) 271-288.
[2] A.R. Feinstein, An analysis of diagnostic reasoning: The strategy of intermediate decisions Yale Biology and medicine, 46 (1973) 264-283.
[3] A.L.Rector and E. Ackerman, Rules for sequential diagnosis, Comput. Bio-Med. Research 8 (1975) 143-155.
[4] Atanassov,K, Intuitionistic fuzzy sets, Fuzzy Sets and Systems, 20 (1986), pp 87-97.
[5] B.S. Duran and T.O. Lesis, An application of cluster analysis to construction of a diagnostic classification. Computers in Biology and Medicine 4 (1974) 183- 188.
[6] C.V. Negoita and D.A. Ralescu, Application of Fuzzy Sets to System Analysis (Wiley. New York, 1976).
[7] Dubious, D. and Prade, H., "Fuzzy Sets and System: Theory and Application", Academic Press, New York . (1990).
[8] Gau, W. L. and Buehrer, D. J, Vague sets, IEEE Transactions on Systems, Man and Cybernetics, Vol.(23), (1993) pp 610-614.
[9] G.A. Gorry and G.O. Barnett, Sequential diagnosis by computer. J. Am. Med. Assoc. 205 (1968) 849-854.
[10] H.R. Warner, A.F. Toronto, L.G. Veasey and R. Stephensio. A mathematical approach to medical diagnosis, Jou.Amer. Med. Assoc. 177 (3) (1961) 177-184.
[11] J.A.Jarquez , Computer Diagnosis and Diagnostic Methods (Thomas, Springfield, IL 1972).
[12] Klir, J.G. and Yuan, B., Fuzzy Sets And Fuzzy Logic: Theory and Applications, Prentice Hall of India, 1995.
[13] M.J. Norusis and J.A. Jecquez, Diagnosis I: Symptom non independence in mathematical models for diagnosis. Comput, Bio-Med, research 8 (1975) 156-172.
[14] M.J. Norusis and J.A. Jacquez, Diagnosis, II: Diagnostic models based on attribute clusters. A proposal and comparisons. Comput, Bio-Med. Research 8 (1975) 173-188.
[15] Winston, P.H., "Artificial Intelligence," Addison Wesley, India, (2000).
[16] Zadeh. L.A., (1965); Fuzzy sets, Infor. and Control (8), pp 338-353.
[17] Z.Pawlak, Rough Sets, International Journal of Info. and Comp.Sc.11 (1982) 341-3$56$.

**First A. Author**
Siddhartha Sankar Biswas is M.Tech in Computer Science from Institute of Technology & management, Gurgaon and is currently working as Lecturer in Gurgaon Institute of Technology & management, Gurgaon.